\documentclass[letterpaper]{article} 
\usepackage{aaai25}  
\usepackage{times}  
\usepackage{helvet}  
\usepackage{courier}  
\usepackage[hyphens]{url}  
\usepackage{graphicx} 
\usepackage{natbib}  
\usepackage{caption} 
\usepackage{amsmath}
\usepackage{booktabs} 
\usepackage{multirow} 
\usepackage{array}    
\usepackage{colortbl} 
\usepackage[table]{xcolor}
\usepackage{amsfonts}  
\usepackage{algorithm}
\usepackage{algorithmic}

\usepackage{newfloat}
\usepackage{listings}
\DeclareCaptionStyle{ruled}{labelfont=normalfont,labelsep=colon,strut=off} 
\floatstyle{ruled}
\newfloat{listing}{tb}{lst}{}
\floatname{listing}{Listing}
\nocopyright 

\title{QD-VMR: Query Debiasing with Contextual Understanding Enhancement\\ for Video Moment Retrieval}
\author{
    Chenghua Gao\textsuperscript{\rm 1,\rm2},
    Min Li\textsuperscript{\rm 1,2}\thanks{Corresponding author.},
    Jianshuo Liu\textsuperscript{\rm 1,\rm2},
    Junxing Ren\textsuperscript{\rm 1,\rm2}\footnotemark[1],
    Lin Chen\textsuperscript{\rm 1,\rm2}\footnotemark[1],\\
    Haoyu Liu\textsuperscript{\rm 3},
    Bo Meng\textsuperscript{\rm 4},
    Jitao Fu\textsuperscript{\rm 1,\rm2},
    Wenwen Su\textsuperscript{\rm 1,\rm2}
}
\affiliations{
    \textsuperscript{\rm 1}Institute of Information Engineering, Chinese Academy of Sciences, Beijing, China\\
    \textsuperscript{\rm 2}School of Cyber Security, University of Chinese Academy of Sciences, Beijing, China\\
    \textsuperscript{\rm 3}Institute of Computing Technology, Chinese Academy of Sciences, Beijing, China\\
    \textsuperscript{\rm 4}First Research Insitute of MPS, Beijing, China

    \{gaochenghua,limin,liujianshuo,renjunxing,chenlin,fujitao,suwenwen\}@iie.ac.cn
}

\usepackage{bibentry}

\begin{document}

\maketitle

\begin{abstract}
Video Moment Retrieval (VMR) aims to retrieve relevant moments of an untrimmed video corresponding to the query. While cross-modal interaction approaches have shown progress in filtering out query-irrelevant information in videos, they assume the precise alignment between the query semantics and the corresponding video moments, potentially overlooking the misunderstanding of the natural language semantics. To address this challenge, we propose a novel model called \textit{QD-VMR}, a query debiasing model with enhanced contextual understanding. Firstly, we leverage a Global Partial Aligner module via video clip and query features alignment and video-query contrastive learning to enhance the cross-modal understanding capabilities of the model. Subsequently, we employ a Query Debiasing Module to obtain debiased query features efficiently, and a Visual Enhancement module to refine the video features related to the query. Finally, we adopt the DETR structure to predict the possible target video moments. Through extensive evaluations of three benchmark datasets, QD-VMR achieves state-of-the-art performance, proving its potential to improve the accuracy of VMR. Further analytical experiments demonstrate the effectiveness of our proposed module. Our code will be released to facilitate future research.
\end{abstract}

\begin{figure}[ht]
\centering
\includegraphics[width=0.45\textwidth]{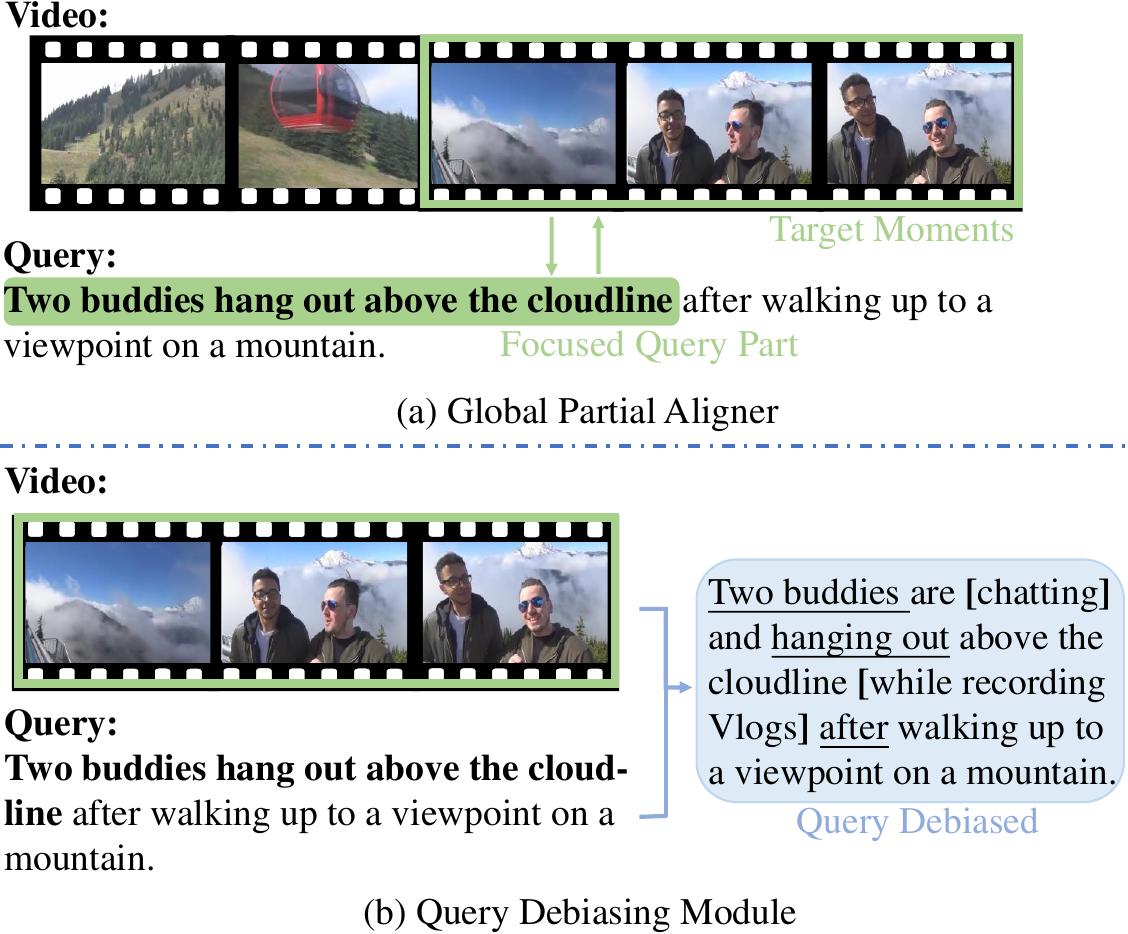} 
\caption{We analyze the query bias issues during VMR. (a) We use the global partial aligner to align the target moments in the video with the parts of the query that need to be focused. (b) We use the query debiasing module to generate the debiased query, with the \underline{underlined} words being enhanced and the \textbf{[}bracketed\textbf{]} parts being expansions.}
\label{fig1}
\end{figure}

\section{Introduction}

Video Moment Retrieval (VMR) is marked as a critical challenge in video understanding, aiming at locating specific moments from an untrimmed video with a given language query \cite{gao2017tall}. Based on DETR \cite{carion2020end}, which leverages a transformer-based model to detect target objects in images, previous studies such as Moment-DETR \cite{lei2021detecting} and QD-DETR \cite{moon2023query} achieve the tasks of VMR and Highlight Detection (HD) by applying well-designed datasets and corresponding annotations. More recently, MESM \cite{liu2024towards} is proposed to better align query-relevant video semantics at both frame-word and segment-sentence levels, significantly refreshing state-of-the-art VMR records.

Although previous studies have made significant efforts to handle irrelevant semantic information in videos, they generally require the entire query semantics as the ground truth to guide the video semantics further, and then predict results. However, due to the rich semantic information of video elements and the model's insufficient and biased contextual understanding of the query, achieving precise video moment retrieval is challenging \cite{moon2023query,liu2024towards,sun2024tr}. No existing work, to our knowledge, has noticed the problem of video moment retrieval with query semantics ambiguous understanding, which brings uncertainty to the retrieval by the model. As shown in Fig. \ref{fig1}(a), given a video and the query ``Two buddies hang out above the cloudline after walking up to a viewpoint on a mountain'', if the model does not understand the context of the sentence and fails to focus on the word ``after'', it will return corresponding video segments related to ``Two buddies hang out above the cloudline'' and ``walking up to a viewpoint on a mountain''. The latter is not the original intent of the query.
Moreover, annotated queries that do not align perfectly with the target video segments worsen the overall performance of retrieval. As illustrated in Fig. \ref{fig1}(b), two buddies in the video are chatting and hanging out while filming vlogs. However, the given query lacks any semantic information related to chatting and vlogging, which may lead to semantic inaccuracies in moment retrieval. 
Solving this issue in VMR tasks requires the model to possess the ability to de-bias the query. Specifically, this includes an accurate understanding of the context of the query and video-query alignment.

In this paper, we propose a novel model named \textit{QD-VMR}, the first query-debiasing video moment retrieval model with contextual understanding enhancement. First, we introduce a Global Partial Aligner (GPA) that aligns video clips with query features and incorporates video-query contrastive learning to enhance the model’s cross-modal understanding capabilities. Then, to tackle the query bias, we propose a Query Debiasing Module (QDM) to acquire debiased query features, and we use a Visual Enhancement (VE) to filter query-irrelevant video features. Finally, we utilize the DETR structure \cite{carion2020end} for result prediction. Extensive experiments on QVHighlights \cite{lei2021detecting}, Charades-STA \cite{gao2017tall} and TACoS \cite{regneri2013grounding} demonstrate that QD-VMR outperforms the SOTA methods. The contributions of this paper are summarized as follows:

\begin{itemize}
\item We propose QD-VMR, which is the first video moment retrieval model that addresses the issue of query semantics ambiguous understanding, thereby improving the accuracy of video moment retrieval.
\item We design a Query Debiasing Module that generates debiased query representations by expanding the query and enhancing contextual understanding. Additionally, we introduce a Global Partial Aligner to ensure alignment between video and text modality, enabling the model to effectively discriminate the most relevant video clips.
\item We show the superiority of the proposed model by extensive experiments; Additionally, we transfer our key modules to a typical VMR approach, with the newly added parameters not exceeding 1M, resulting in significant improvements in all evaluation metrics.
\end{itemize}

\section{Related Work}

\textbf{Video Moment Retrieval.} Video Moment Retrieval (VMR) is first proposed by TALL \cite{gao2017tall}, aiming to locate specific moments from an untrimmed video with a given language query. Various methods have been developed to tackle this task. 2D-TAN \cite{zhang2020learning} covers diverse video moments with different lengths while representing their adjacent relations. Furthermore, with the advent of Moment-DETR \cite{lei2021detecting}, the VMR task has evolved into a new end-to-end pipeline. First, the pre-trained models such as CLIP \cite{radford2021learning} and SlowFast \cite{chen2019semantic} are developed to extract features from both the video and the query text separately \cite{li2020hero}, thereby fusing the extracted features. Subsequently, a DETR-based \cite{carion2020end} model is further proposed to predict the start and end times of the video moments. In recent years, models have emerged to focus more on exploring more suitable representations of video, query, and their alignment\cite{li2023progressive}. QD-DETR \cite{moon2023query} introduces a query-dependent video representation module, making moment predictions reliant on user queries. 
MH-DETR \cite{xu2023mh} introduces a pooling operation into the encoder and incorporates a cross-modality interaction module to fuse visual and query features. 
BM-DETR \cite{jung2023overcoming} adopts a contrastive approach, carefully utilizing the negative queries matched to other moments in the video. CG-DETR \cite{moon2023correlation} leverages an adaptive cross-attention layer with dummy tokens, preventing irrelevant video clips from being represented by the text query. TR-DETR \cite{sun2024tr} fully exploits the reciprocal relationship between VMR and HD. MESM \cite{liu2024towards} achieves more balanced semantic alignment at both the frame-level and segment-sentence level. However, most DETR-based methods mainly address redundant features in the video modality while neglecting issues with the query.\\
\textbf{Query Bias and Partial Relevance} In the area of information retrieval, the following two issues are widely discussed. Query-Document Mismatching \cite{li2014semantic, liu-etal-2024-se2} occurs when the information needs expressed by the user in the query is not well-aligned with the content or context of the documents retrieved. This can happen due to vocabulary differences, context misunderstandings, or ambiguous queries. Partial Relevance \cite{hou2024improving} refers to a situation where the document partially meets the user's information need but is not wholly focused on the topic. These two issues persist in video moment retrieval, which we refer to as query bias and partial relevance problems. Currently, several methods provide solutions to the issue of partial relevance between query and video. Cross-modal interaction \cite{lei2021detecting,xu2023mh,sun2024tr,hou2024improving} is widely used to reconstruct video features to obtain parts relevant to the query. However, the problem of query bias remains to be explored. To address this issue, we enhance the model's contextual understanding of the query and video-query alignment by expanding the query features and using the combined video and query features to jointly predict the masked words.

\begin{figure*}[ht]
\centering
\includegraphics[width=0.9\textwidth]{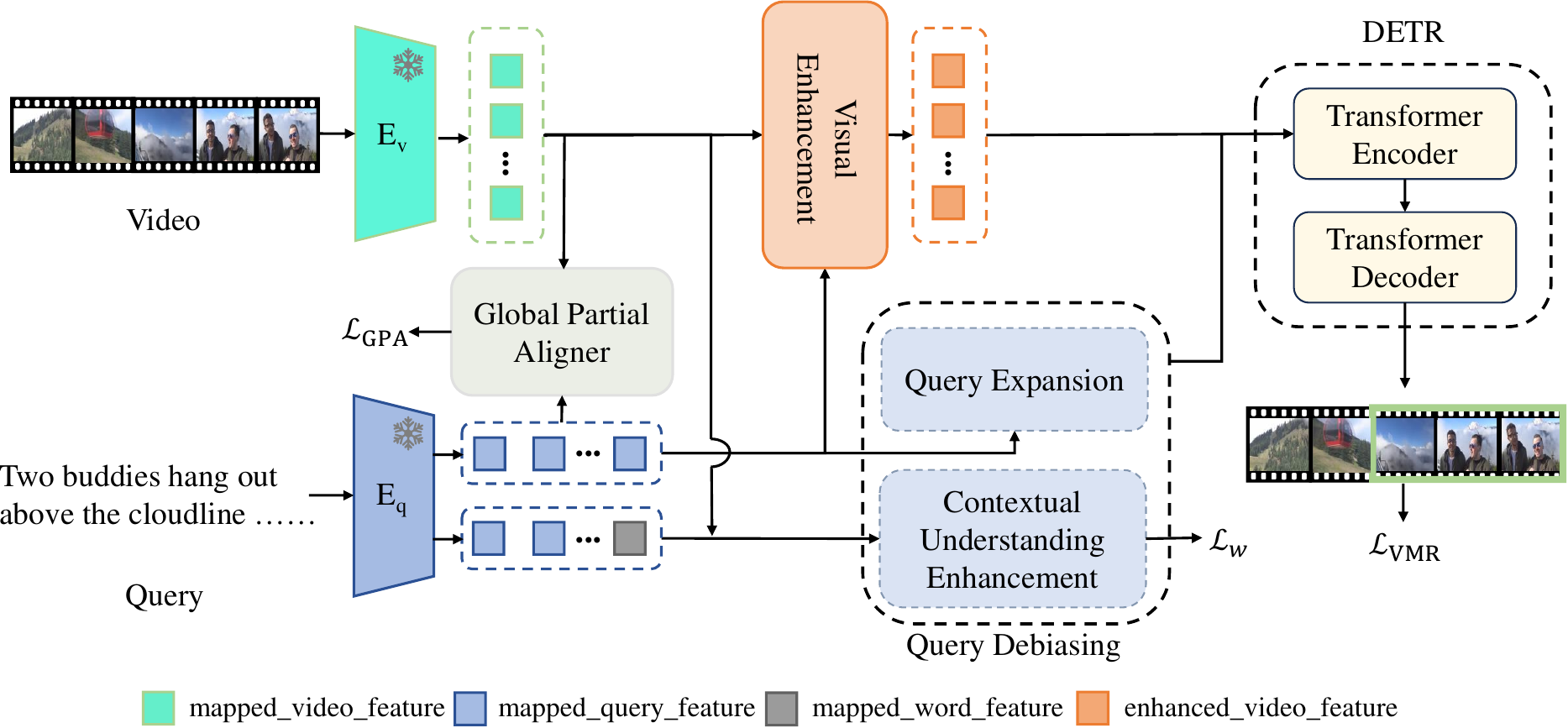} 
\caption{An overview of QD-VMR. First, the video and query are processed through pre-trained feature extractors to obtain the corresponding features. After being mapped to the same space, a Global Partial Aligner is adopted to align the video and query features. Then, a Query Debiasing Module and a Visual Enhancement are employed to de-bias the query features and enhance the video features related to the query, respectively. Finally, a DETR structure is implemented for result prediction. The Query Debiasing Module consists of two parts: Query Expansion and Contextual Understanding Enhancement. Additionally, the encoder in the Query Expansion and the encoder in the DETR share weights.}
\label{fig2}
\end{figure*}

\section{Method}
\subsection{Overview}
Video Moment Retrieval (VMR) aims to identify the most relevant moments from a video, based on the user's natural language query. Since existing methods may ignore the possible semantic bias in query texts, QD-VMR is specially designed to fill in the gap between insufficient contextual understanding of query and targeted video content. We represent the video and text features extracted by the pre-trained models as \(F_v\) and \(F_t\). After mapping features into the same vector space, we adopt the Global Partial Aligner to get better alignment between video and query. Then, we leverage Query Debiasing and Visual Enhancement modules to further improve the quality of query and video features. Finally, a DETR-based network is utilized to encode the concatenated query and video features, and then decode the main objective, which is used to localize the center coordinate \(m_c\) and corresponding duration \(m_\sigma\) \cite{moon2023correlation}. The overall architecture of QD-VMR is shown in Fig. \ref{fig2}, and we present the detailed design of the scheme as follows.

\subsection{Feature Extraction}
We follow previous works \cite{li2020hero, lei2021detecting, sun2024tr} by leveraging pre-trained networks to extract raw visual and textual features as inputs. To enhance the model's ability to understand the semantic context of queries, we additionally use the Mask Language Model (MLM) to randomly mask one-third of the words in the query. For both complete queries and partially masked queries, we adopt the CLIP's text encoder module \cite{radford2021learning, liu2024towards} to extract complete query features \(
F_t = \left\{ f_t^i  \in \mathbb{R}^{N \times D_t} \mid i = 1, \ldots, K \right\}
\), and partially-masked query features \(
F_w = \left\{ f_w^i  \in \mathbb{R}^{N \times D_t} \mid i = 1, \ldots, K \right\}
\), where \(K\) is the pairs of video and query in the dataset, \(N\) and \(D_t\) are the length and dimension of textual features, respectively. Moreover, we extract clip-level features from the video given. Specifically, we first divide the video into non-overlapping clips of shorter duration (e.g., 2 seconds) and extract the clip-level video features by using pre-trained SlowFast \cite{feichtenhofer2019slowfast} and CLIP's visual encoder \cite{radford2021learning}. Then, both Slowfast and CLIP features are concatenated as the overall video features \(
F_v = \left\{ f_v^i  \in \mathbb{R}^{L \times D_v} \mid i = 1, \ldots, K \right\}
\), where \(L\) and \(D_v\) are the length and dimension of visual features, respectively. Following the approach in \cite{liu2022umt,sun2024tr}, we utilize a pre-trained audio feature extractor to extract audio features from the video, and subsequently concatenate the extracted audio and video features. More details are provided in the experiment section. Furthermore, we use trainable MLPs to map the visual and textual features to the same feature space, resulting in:
\begin{equation}
    \overline{F}_k = \text{MLP}_k(F_k),
\end{equation}
where \(k \) can be replaced by $v$, $t$ and $w$, respectively; $\overline{F}_v$ represents mapped video features \(\overline{F}_v = \{\overline{f}_v^i  \in \mathbb{R}^{L \times H} \mid i = 1, \ldots, K \}\), 
and both \(\overline{F}_t = \{ \overline{f}_t^i  \in \mathbb{R}^{N \times H} \mid i = 1, \ldots, K \}\) and \(\overline{F}_w = \{\overline{f}_w^i  \in \mathbb{R}^{N \times H} \mid i = 1, \ldots, K \}\) altogether delineate mapped text features; $H$ is dimension of hidden layers.

\subsection{Global Partial Aligner}
After mapping the encoded video and text features into the same vector space, the current method \cite{liu2024towards} directly fuses both mapped features to enhance the parts of the video features related to the query. However, even though both video and text features encoded by Slowfast \cite{feichtenhofer2019slowfast} and CLIP \cite{carion2020end} have the potential to unveil the aligned semantic information of images with single frame, the temporal dimension gap between video and text features still exists. Therefore, by analyzing the fused features, the gap may eventually render a poor understanding of temporal boundaries and events within the boundaries \cite{huang2024vtimellm}. To achieve better spatiotemporal alignment of features and perception of event boundaries, we leverage the idea from \cite{sun2024tr} and further design the Global Partial Aligner. 
Given the visual features \(\overline{F}_v\) and textual features \(\overline{F}_t\), the similarity between both features can be obtained as:
\begin{equation}
    S = \frac{\overline{F}_t \overline{F}_v^\top}{\|\overline{F}_t\| \|\overline{F}_v\|}.
\end{equation}
Then, mean-pooling is applied in the last dimension to get $\overline{S}$ which measures the similarity between video clips and the entire query features. We use a Part-aware loss $\mathcal{L}_\text{P}$ to enhance the model's ability to discriminate the temporal boundaries of events for paired video and query.
\begin{equation}
    \mathcal{L}_{\text{P}} = -\sum_{i=1}^{L} \left( C_i \log(\overline{S}_i) + (1 - C_i) \log(1 - \overline{S}_i) \right),
\end{equation}
where \(\overline{S}_i\) is the similarity score between the \(i\)-th video clip and entire query features, and \(C_i \in \{0, 1\}\) indicates whether the video clip is relevant to the query in the ground truth. In addition, we use the InfoNCE loss function to make the model align the semantic representations of the paired videos and queries:
\begin{equation}
    \mathcal{L}_\text{G} = -\frac{1}{B} \sum_{i=1}^{B} \log \frac{\exp((\overline{F}^{Gi}_v)(\overline{F}^{Gi}_t)^\top / \tau)}{\sum_{j=1}^{B} \exp((\overline{F}^{Gi}_v)(\overline{F}^{Gj}_t)^\top / \tau)},
\end{equation}
where \(\overline{F}^{Gi}_v\) denotes the average of the clip features for the \(i\)-th video in a batch $B$, \(\overline{F}^{Gi}_t\) is the average of the word features for the \(i\)-th query, $\tau \in (0,1]$ is the temperature parameter. We calculate the average of \(\mathcal{L}_{\text{P}}\) and \(\mathcal{L}_\text{G}\) to obtain \(\mathcal{L}_{\text{GPA}}\).

\subsection{Query Debiasing Module}
Existing methods primarily focus on partial relevance issues within the video. In addition, accurate and complete query semantics are crucial for video moment retrieval. Unfortunately, existing works do not carefully consider query bias among the query texts. To address this issue, we propose a query debiasing module that consists of two main parts: query expansion and contextual understanding enhancement. Query expansion is designed to learn video semantic information not included in the query through relevant video segments.
Assume the hyperparameter $N_e$, and the learnable parameter $F_e=\left\{f_e^i \in \mathbb{R}^{N_e \times H}  \mid i = 1, \ldots, K \right\}$, the expanded text features $F'_e$ can be further expressed as:
\begin{equation}
    F'_e=\text{encoder}(F_e||\overline{F}_t),
\end{equation}
where ``\(||\)'' denotes concatenation operator, ``encoder'' refers to the Transformer Encoder shown in DETR scheme. Moreover, we extract top \(N_e\) features from \(F'_e\) and concatenate them with \(\overline{F}_t\) to obtain the expanded text features \(\hat{F}_t\).

Furthermore, simply using query expansion to enrich the information in the query is not sufficient, as there is often a mismatching between the information expressed by the user in the query and the content or context of the video. For the partially masked queries obtained in the Feature Extraction subsection, we use contextual understanding enhancement to further mitigate the bias between the query and the video. It enhances the model's awareness of keywords by predicting them in the query based on the video and query context features within the global partial aligner. The enhanced feature of words $F_w$ can be calculated as:
\begin{equation}
    F_w^{R} = \overline{F}_w + \text{MLP} \left( \text{softmax} \left( \frac{\mathcal{QK}^\top}{\sqrt{d}} \right) \mathcal{V} \right),
\end{equation}
where $\mathcal{Q}=\overline{F}_v$ as \textit{query}, $\mathcal{K}=\mathcal{V}=\overline{F}_w$ as \textit{key} and \textit{value}. 
Then we leverage MLP and softmax to calculate the probability distribution of the masked words as $P(F_w^R) \in \mathbb{R}^{N \times L_{\text{vocab}}}$, where $L_{\text{vocab}}$ is the vocabulary size. Moreover, the contextual understanding enhancement loss function is expressed as:
\begin{equation}
\mathcal{L}_{w} = - \frac{1}{N} \sum_{i=1}^{N} \log P(F_w^R).
\end{equation}

\subsection{Visual Enhancement}
Fused video-text features play an essential role in enhancing the prediction of temporal boundaries for semantic features in the video, which is relevant to the query text. Moment-DETR \cite{lei2021detecting} simply concatenates visual and textual features for modal interaction. MESM \cite{liu2024towards} enhances the video modality at the frame-word level by emphasizing the query-relevant portions of frame-level features and suppressing irrelevant ones. Following TR-DETR \cite{sun2024tr}, we use visual enhancement to fuse the video and text modalities. First, we perform row softmax normalization and column softmax normalization on the similarity matrix \( S \) to obtain the results \( S_r \) and \( S_c \), and then compute \( F_{v2q} \) and \( F_{q2v} \), which are the clip-level textual features and word-level visual features, respectively.
\begin{equation}
    F_{v2q}=S_r\overline{F}_t,
\end{equation}
\begin{equation}
    F_{q2v}=S_rS_c^\top\overline{F}_t.
\end{equation}
Finally, we obtain the enhanced video features $\hat{F}_v$ by using the following feature concatenation and linear projection:
\begin{equation}
    \hat{F}_v = \text{Linear}\left[\overline{F}_v||F_{v2q}||\overline{F}_v \odot F_{v2q}||\overline{F}_v \odot F_{q2v}||\overline{F}_t^G\right].
\end{equation}

\begin{table*}[t]
\centering
\begin{tabular}{lllllllll}
\toprule
&  & \multicolumn{5}{c}{\textbf{Moment Retrieval}} & \multicolumn{2}{c}{\textbf{HD}} \\
\cmidrule(lr){3-7}
\cmidrule(lr){8-9}
\textbf{Method} & \multicolumn{1}{c}{\textbf{Src}} & \multicolumn{2}{c}{R1} & \multicolumn{3}{c}{mAP} & \multicolumn{2}{c}{\(\geq\) Very Good} \\
\cmidrule(lr){3-4}
\cmidrule(lr){5-7}
\cmidrule(lr){8-9}
&  & @0.5 & @0.7 & @0.5 & @0.75 & Avg. & mAP & HIT@1 \\
\midrule
MCN \cite{anne2017localizing} & \multicolumn{1}{c}{V} & 11.41 & 2.72 & 24.94 & 8.22 & 10.67 & \multicolumn{1}{c}{-} & \multicolumn{1}{c}{-}\\
CAL \cite{escorcia2019temporal} & \multicolumn{1}{c}{V} & 25.49 & 11.54 & 23.40 & 7.65 & \multicolumn{1}{c}{9.89} & \multicolumn{1}{c}{-} & \multicolumn{1}{c}{-}\\
XML \cite{lei2020tvr} & \multicolumn{1}{c}{V} & 41.83 & 30.35 & 44.63 & 31.73 & 32.14 & 34.49 & \multicolumn{1}{c}{55.25}\\
XML+ \cite{lei2020tvr} & \multicolumn{1}{c}{V} & 46.69 & 33.46 & 47.89 & 34.67 & 34.90 & 35.38 & \multicolumn{1}{c}{55.06}\\
M-DETR \cite{lei2021detecting} & \multicolumn{1}{c}{V} & 52.89 & 33.02 & 54.82 & 29.40 & 30.73 & 35.69 & \multicolumn{1}{c}{55.60}\\
UniVTG \cite{lin2023univtg} & \multicolumn{1}{c}{V} & 58.86 & 40.86 & 57.60 & 35.59 & 35.47 & 38.20 & \multicolumn{1}{c}{60.96}\\
QD-DETR \cite{moon2023query} & \multicolumn{1}{c}{V} & 62.40 & 44.98 & 62.62 & 39.88 & 39.86 & 38.64 & \multicolumn{1}{c}{62.40}\\
QD-DETR* \cite{moon2023query} & \multicolumn{1}{c}{V} & \underline{65.30} & \underline{46.56} & \underline{64.50} & \underline{40.94} & \underline{41.44} & \underline{39.66} & \multicolumn{1}{c}{\underline{63.75}}\\
CG-DETR \cite{moon2023correlation} & \multicolumn{1}{c}{V} & 65.43 & 48.38 & 64.51 & 42.77 & 42.86 & \multicolumn{1}{c}{40.33} & \multicolumn{1}{c}{\textbf{66.21}} \\
MESM \cite{liu2024towards} & \multicolumn{1}{c}{V} & 62.78 & 45.20 & 62.64 & 41.45 & 40.68 & \multicolumn{1}{c}{-} & \multicolumn{1}{c}{-} \\
TR-DETR \cite{sun2024tr} & \multicolumn{1}{c}{V} & 64.66 & \textbf{48.96} & 63.98 & \textbf{43.73} & 42.62 & 39.91 & \multicolumn{1}{c}{63.42} \\
\rowcolor{lightgray}
QD-VMR(Ours) & \multicolumn{1}{c}{V} & \textbf{66.73} & 48.64 & \textbf{66.04} & 42.80 & \textbf{43.06} & \multicolumn{1}{c}{\textbf{40.50}} & \multicolumn{1}{c}{65.76} \\
\midrule
UMT \cite{liu2022umt} & V+A & 56.23 & 41.18 & 53.38 & 37.01 & 36.12 & 38.18 & \multicolumn{1}{c}{59.99}\\
QD-DETR \cite{moon2023query} & V+A & 63.06 & 45.10 & 63.04 & 40.10 & 40.19 & 39.04 & \multicolumn{1}{c}{62.87}\\
TR-DETR \cite{sun2024tr} & V+A & 65.05 & 47.67 & 64.87 & 42.98 & 43.10 & 39.90 & \multicolumn{1}{c}{63.88} \\
\rowcolor{lightgray}
QD-VMR(Ours) & V+A & \textbf{66.93} & \textbf{50.13} & \textbf{66.60} & \textbf{44.39} & \textbf{44.61} & \multicolumn{1}{c}{\textbf{40.43}} & \multicolumn{1}{c}{\textbf{65.30}} \\
\bottomrule
\end{tabular}
\caption{Performance comparison (\%) on Qvhighlights \textit{test} set. ``*'' denotes the result we re-implement QD-DETR with our key modules under the same training scheme, and metrics with ``\underline{ }'' indicate the improved performance. All the listed methods use SlowFast+CLIP as their video extractor and CLIP as their text extractor. M-DETR is short for Moment-DETR.}
\label{table1}
\end{table*}

\begin{table}[t]
\centering
\small
\begin{tabular}{llll}
\toprule
\textbf{Method} & \multicolumn{1}{c}{\textbf{Feat}} & R1@0.5 & R1@0.7 \\
\midrule
2D-TAN \cite{zhang2020learning} & SF+C & \multicolumn{1}{c}{46.02} & \multicolumn{1}{c}{27.50} \\
VSLNet \cite{zhang2020span} & SF+C & \multicolumn{1}{c}{42.69} & \multicolumn{1}{c}{24.14} \\
LLaViLo \cite{ma2023llavilo} & SF+C & \multicolumn{1}{c}{55.72} & \multicolumn{1}{c}{33.43} \\
UniVTG \cite{lin2023univtg} & SF+C & \multicolumn{1}{c}{58.01} & \multicolumn{1}{c}{35.65}\\
CG-DETR \cite{moon2023correlation} & SF+C & \multicolumn{1}{c}{58.44} & \multicolumn{1}{c}{36.34} \\
QD-DETR \cite{moon2023query} & SF+C & \multicolumn{1}{c}{57.31} & \multicolumn{1}{c}{32.55} \\
MESM \cite{liu2024towards} & SF+C & \multicolumn{1}{c}{\textbf{61.24}} & \multicolumn{1}{c}{\textbf{38.04}} \\
TR-DETR \cite{sun2024tr} & SF+C& \multicolumn{1}{c}{57.61} & \multicolumn{1}{c}{33.52} \\
\rowcolor{lightgray}
QD-VMR(Ours) & SF+C & \multicolumn{1}{c}{58.55} & \multicolumn{1}{c}{36.34} \\
\midrule
SAP \cite{chen2019semantic} & VGG & \multicolumn{1}{c}{27.42} & \multicolumn{1}{c}{13.36} \\
MAN \cite{zhang2019man} & VGG & \multicolumn{1}{c}{41.24} & \multicolumn{1}{c}{20.54} \\
TripNet \cite{hahn2019tripping} & VGG & \multicolumn{1}{c}{36.61} & \multicolumn{1}{c}{14.50} \\
2D-TAN \cite{zhang2020learning} & VGG & \multicolumn{1}{c}{40.94} & \multicolumn{1}{c}{22.85} \\
FVMR \cite{gao2021fast} & VGG & \multicolumn{1}{c}{42.36} & \multicolumn{1}{c}{24.14} \\
UMT* \cite{liu2022umt} & VGG & \multicolumn{1}{c}{48.31} & \multicolumn{1}{c}{29.25} \\
QD-DETR \cite{moon2023query} & VGG & \multicolumn{1}{c}{52.77} & \multicolumn{1}{c}{31.13} \\
QD-DETR* \cite{moon2023query} & VGG & \multicolumn{1}{c}{55.51} & \multicolumn{1}{c}{34.17} \\
CG-DETR \cite{moon2023correlation} & VGG & \multicolumn{1}{c}{55.22} & \multicolumn{1}{c}{34.19} \\
TR-DETR \cite{sun2024tr} & VGG & \multicolumn{1}{c}{53.47} & \multicolumn{1}{c}{30.81}\\
TR-DETR* \cite{sun2024tr} & VGG & \multicolumn{1}{c}{54.49} & \multicolumn{1}{c}{32.37}\\
\rowcolor{lightgray}
QD-VMR(Ours) & VGG & \multicolumn{1}{c}{\textbf{56.91}} & \multicolumn{1}{c}{\textbf{35.65}} \\
\bottomrule
\end{tabular}
\caption{Performance comparison (\%) on Charades-STA \textit{test} set. ``*'' represents using audio features. All the listed methods use CLIP as their text extractor.}
\label{table2}
\end{table}

\begin{table}[t]
\centering
\small
\begin{tabular}{lll}
\toprule
\textbf{Method} & R1@0.5 & R1@0.7 \\
\midrule
2D-TAN \cite{zhang2020learning} & \multicolumn{1}{c}{27.99} & \multicolumn{1}{c}{12.92} \\
VSLNet \cite{zhang2020span} & \multicolumn{1}{c}{23.54} & \multicolumn{1}{c}{13.15} \\
M-DETR \cite{lei2021detecting} & \multicolumn{1}{c}{24.67} & \multicolumn{1}{c}{11.97} \\
UniVTG \cite{lin2023univtg} & \multicolumn{1}{c}{34.97} & \multicolumn{1}{c}{17.35} \\
CG-DETR \cite{moon2023correlation} & \multicolumn{1}{c}{\textbf{39.61}} & \multicolumn{1}{c}{\textbf{22.23}} \\
\rowcolor{lightgray}
QD-VMR(Ours) & \multicolumn{1}{c}{32.67} & \multicolumn{1}{c}{20.64} \\
\bottomrule
\end{tabular}
\caption{Performance comparison (\%) on TACoS. All the listed methods use SlowFast+CLIP as their video extractor and CLIP as their text extractor. M-DETR is short for Moment-DETR.}
\label{table3}
\end{table}

\subsection{Transformer Encoder-Decoder}
After concatenating the debiased text features $\hat{F}_t$ and the enhanced video features $\hat{F}_v$ to obtain the modality-aligned feature \(F\), we adopt a DETR-structure module to predict the target video moments.
It consists of a transformer encoder and a decoder. Specifically, the encoder is utilized to fuse the features \(F\) from the two modalities. In designing the decoder, we follow the approach of \cite{moon2023query,liu2024towards}, incorporating learnable spans that represent the center coordinate \(m_c\) and the duration \(m_\sigma\).

\subsection{Objective Losses}
Following previous studies \cite{carion2020end,lei2021detecting, liu2024towards}, we design the moment retrieval loss $\mathcal{L}_{\text{VMR}}$, which consists of three parts:
\begin{equation}
    \mathcal{L}_{\text{VMR}} = \lambda_{L_1} \| m - \hat{m} \|_1 + \lambda_{\text{iou}} \mathcal{L}_{\text{iou}}(m, \hat{m}) + \lambda_{\text{ce}} \mathcal{L}_{\text{ce}},
\end{equation}
where $m$ and $\hat{m}$ denote the prediction and ground-truth, respectively; $ \lambda_{L1},\lambda_{\text{iou}}, \lambda_{\text{ce}}$ are the hyper-parameters; $\mathcal{L}_{\text{iou}}$ is the GIoU loss computed by the predicted span and target span, $\mathcal{L}_{\text{ce}}$ refers to the cross-entropy loss, which is used to classify the foreground or background \cite{carion2020end}.
As a result, the final loss is calculated as:
\begin{equation}
\mathcal{L}=\lambda_{\text{GPA}}\mathcal{L}_{\text{GPA}}+\lambda_w\mathcal{L}_w+\mathcal{L}_{\text{VMR}},
\end{equation}
where $\lambda_{\text{GPA}}$ and $\lambda_w$ are the hyper-parameters.

\begin{table*}[t]
\centering
\begin{tabular}{llllllllllll}
\toprule
 & \multicolumn{4}{c}{\textbf{Module}} & \multicolumn{5}{c}{\textbf{Moment Retrieval}} & \multicolumn{2}{c}{\textbf{HD}} \\
\cmidrule(lr){2-5}
\cmidrule(lr){6-10}
\cmidrule(lr){11-12}
\multicolumn{1}{c}{\textbf{Setting}} & \multirow{2}[1]{*}{GPA} & \multirow{2}[1]{*}{VE} & \multicolumn{2}{c}{QDM} & \multicolumn{2}{c}{R1} & \multicolumn{3}{c}{mAP} & \multicolumn{2}{c}{\(\geq\) Very Good} \\
\cmidrule(lr){4-5}
\cmidrule(lr){6-7}
\cmidrule(lr){8-10}
\cmidrule(lr){11-12}
 & && \multicolumn{1}{c}{QE} & \multicolumn{1}{c}{CUE} &@0.5 & @0.7 & @0.5 & @0.75 & Avg. & mAP & HIT@1 \\
\midrule
 \multicolumn{1}{c}{(a)} & &&&& \multicolumn{1}{c}{59.74} & \multicolumn{1}{c}{45.42} & \multicolumn{1}{c}{59.79} & \multicolumn{1}{c}{41.30} & \multicolumn{1}{c}{40.45} & \multicolumn{1}{c}{37.78} & \multicolumn{1}{c}{60.26}\\
\midrule
 \multicolumn{1}{c}{(b)} & \multicolumn{1}{c}{\checkmark} &&&& \multicolumn{1}{c}{63.61} & \multicolumn{1}{c}{48.65} & \multicolumn{1}{c}{62.94} & \multicolumn{1}{c}{43.90} & \multicolumn{1}{c}{42.22} & \multicolumn{1}{c}{39.71} & \multicolumn{1}{c}{63.68}\\
 \multicolumn{1}{c}{(c)} & & \multicolumn{1}{c}{\checkmark} &&& \multicolumn{1}{c}{63.48} & \multicolumn{1}{c}{48.00} & \multicolumn{1}{c}{62.87} & \multicolumn{1}{c}{43.01} & \multicolumn{1}{c}{42.03} & \multicolumn{1}{c}{39.44} & \multicolumn{1}{c}{62.71}\\
 \multicolumn{1}{c}{(d)} & & & \multicolumn{1}{c}{\checkmark} && \multicolumn{1}{c}{61.68} & \multicolumn{1}{c}{48.52} & \multicolumn{1}{c}{62.13} & \multicolumn{1}{c}{43.04} & \multicolumn{1}{c}{42.90} & \multicolumn{1}{c}{38.28} & \multicolumn{1}{c}{59.74}\\
 \multicolumn{1}{c}{(e)} & & & & \multicolumn{1}{c}{\checkmark} & \multicolumn{1}{c}{59.35} & \multicolumn{1}{c}{45.29} & \multicolumn{1}{c}{60.03} & \multicolumn{1}{c}{41.50} & \multicolumn{1}{c}{40.67} & \multicolumn{1}{c}{38.00} & \multicolumn{1}{c}{58.52}\\
 \multicolumn{1}{c}{(f)} & & & \multicolumn{1}{c}{\checkmark} & \multicolumn{1}{c}{\checkmark} & \multicolumn{1}{c}{62.00} & \multicolumn{1}{c}{47.16} & \multicolumn{1}{c}{62.74} & \multicolumn{1}{c}{43.36} & \multicolumn{1}{c}{42.41} & \multicolumn{1}{c}{38.75} & \multicolumn{1}{c}{60.84}\\
\midrule
 \multicolumn{1}{c}{(g)} & \multicolumn{1}{c}{\checkmark} & \multicolumn{1}{c}{\checkmark} &&& \multicolumn{1}{c}{67.16} & \multicolumn{1}{c}{50.84} & \multicolumn{1}{c}{66.34} & \multicolumn{1}{c}{45.46} & \multicolumn{1}{c}{44.85} & \multicolumn{1}{c}{41.10} & \multicolumn{1}{c}{66.45}\\
 \multicolumn{1}{c}{(h)} & \multicolumn{1}{c}{\checkmark} & & \multicolumn{1}{c}{\checkmark} & \multicolumn{1}{c}{\checkmark} & \multicolumn{1}{c}{64.90} & \multicolumn{1}{c}{50.71} & \multicolumn{1}{c}{64.38} & \multicolumn{1}{c}{45.14} & \multicolumn{1}{c}{44.69} & \multicolumn{1}{c}{40.78} & \multicolumn{1}{c}{65.87} \\
 \multicolumn{1}{c}{(i)} & & \multicolumn{1}{c}{\checkmark} & \multicolumn{1}{c}{\checkmark} & \multicolumn{1}{c}{\checkmark}  & \multicolumn{1}{c}{62.13} & \multicolumn{1}{c}{46.13} & \multicolumn{1}{c}{62.25} & \multicolumn{1}{c}{41.56} & \multicolumn{1}{c}{40.89} & \multicolumn{1}{c}{38.51} & \multicolumn{1}{c}{60.65}\\
\midrule
\multicolumn{1}{c}{(j)} & \multicolumn{1}{c}{\checkmark} & \multicolumn{1}{c}{\checkmark} & \multicolumn{1}{c}{\checkmark} & \multicolumn{1}{c}{\checkmark} & \textbf{67.68} & \textbf{52.32} & \textbf{67.10} & \textbf{48.14} & \textbf{46.18} & \multicolumn{1}{c}{\textbf{41.61}} & \multicolumn{1}{c}{\textbf{67.35}} \\
\bottomrule
\end{tabular}
\caption{Comparison (\%) with the baseline (TR-DETR \cite{sun2024tr} without LGAM and VFR) with different module combinations on QVHighlights \textit{val} set. GPA represents the global partial aligner, VE is the visual enhancement, and QDM represents the query debiasing module which consists of query expansion and contextual understanding enhancement.}
\label{table4}
\end{table*}

\begin{figure*}[ht]
\centering
\includegraphics[width=1\textwidth]{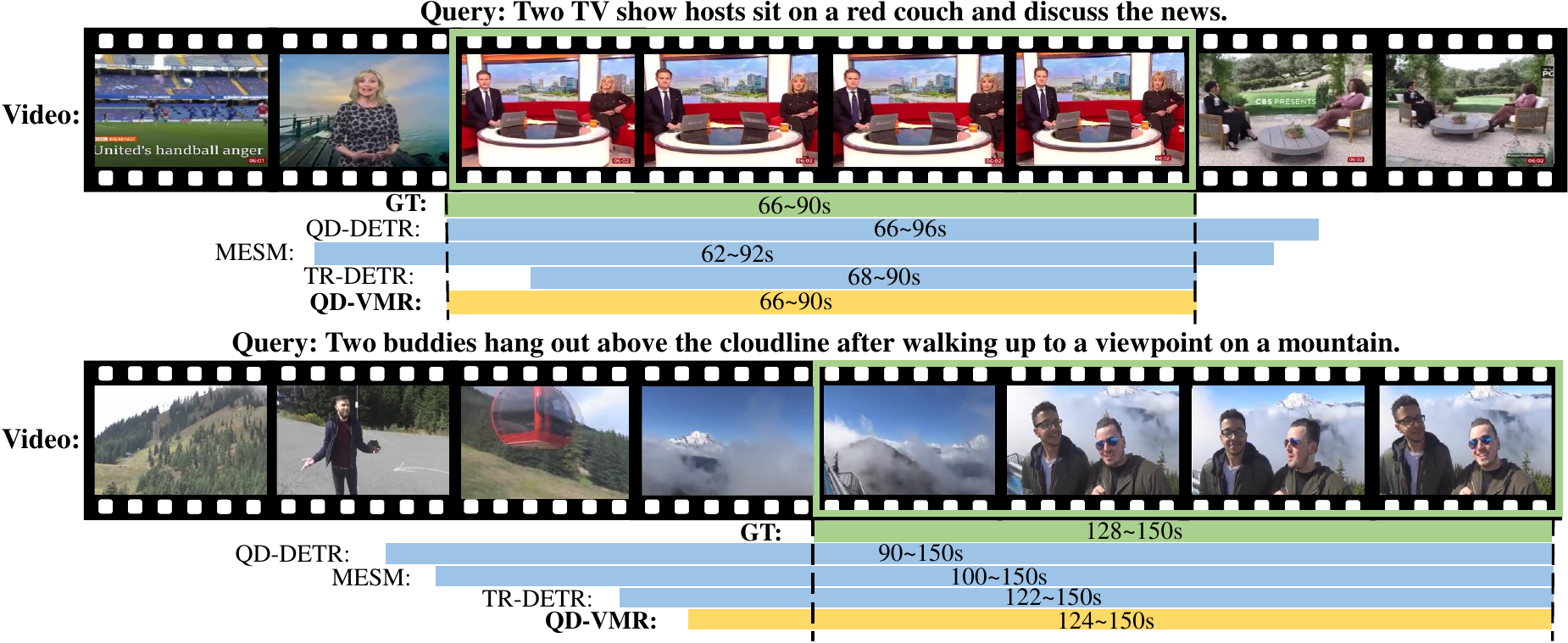} 
\caption{Visualization on Qvighlights \textit{val} split. QD-VMR can have a better understanding of the key information in the query.}
\label{fig3}
\end{figure*}

\section{Experiments}

\subsection{Experimental Settings}
\textbf{Datasets.} We evaluate the proposed method on three widely used datasets. The QVHighlights dataset \cite{lei2021detecting} contains over 10,000 YouTube videos covering a wide range of topics, from daily life to news events. Performance on the test split is rigorously evaluated by uploading the results to the Codalab platform. 
The Charades-STA dataset \cite{gao2017tall} primarily consists of videos focused on indoor activities. Following \cite{moon2023query,sun2024tr}, we use 12,408 samples for training and reserve the remaining 3,720 samples for testing. The TACoS dataset \cite{regneri2013grounding} is specifically centered around cooking-related activities.\\
\textbf{Metrics and Implementation Details.} We use standard metrics from recent studies \cite{lei2021detecting,moon2023query,moon2023correlation,liu2024towards,sun2024tr} to evaluate performance. For QVHighlights, we calculate Recall@1 with IoU thresholds of ${0.5, 0.7}$ and mAP with IoU thresholds of ${0.5, 0.75}$. Additionally, we uniformly sample 10 IoU thresholds from ${0.5, 0.95}$ to compute mAP and average these values as the final mAP metric for the VMR task. We also calculate mAP and HIT@1 for the HD task. For Charades-STA and TACoS, we use Recall@1 with IoU thresholds of ${0.5, 0.7}$.

To validate the generality of our method, we conducted experiments using different pre-trained video feature extractors, with and without the inclusion of audio features as variables. Specifically, for QVHighlights, we used SlowFast+CLIP to extract video features, CLIP for text features, and PANN \cite{kong2020panns} for audio features. The training parameters were set to 200 epochs, a batch size of 256, and a learning rate of 2e-4. For Charades-STA, we used either SlowFast+CLIP or VGG to extract video features, and CLIP for text features, with training set to 200 epochs, a batch size of 32, and a learning rate of 2e-4. For TACoS, we used SlowFast+CLIP for video features and CLIP for text features, with the same hyperparameters as for Charades-STA. All experiments were conducted on Nvidia RTX 3090, A800 GPUs, and an Intel(R) Xeon(R) Gold 6226R CPU.

\subsection{Performance Comparisons}
Tables \ref{table1}, \ref{table2}, and \ref{table3} respectively present the performance comparison of our method against other leading approaches on the three datasets.

In Table \ref{table1}, we compare the performance on the QVHighlights \textit{test} set. We extract the CLIP+Slowfast features for the video and the CLIP features for the text. In the Src column, ``V'' indicates the use of video features only, while ``V+A'' indicates the use of both video and audio features. The results indicate that our method achieves the best performance on most evaluation metrics without using audio data. When audio data is incorporated, our method achieves the best performance across all evaluation metrics. This demonstrates that our method, based on understanding the query context, can better comprehend the combined features of video and audio. To further validate the effectiveness of our approach, we incorporated the key modules of our proposed method into QD-DETR \cite{moon2023query}, a popular and widely used baseline. The results show that our method comprehensively improves the performance of QD-DETR with an increase of less than 1M parameters, particularly enhancing R1@0.5 by 2.9\%. 

Table \ref{table2} presents the performance comparison of QD-VMR on the Charades-STA \textit{test} set. We use Slowfast+CLIP and VGG as video feature extractors, and CLIP as the text feature extractor. The results demonstrate that our model also performs exceptionally well on the Charades-STA dataset. Particularly, when using VGG video features and CLIP text features, QD-VMR outperforms the current state-of-the-art methods across all metrics.

We also show a performance comparison of the TACoS dataset in Table 3. Interestingly, we find that QD-VMR performs less satisfactorily on this dataset. We speculate that this is due to the video content in the TACoS dataset being limited to the cooking domain, where QD-VMR's ability to capture subtle differences in actions and objects is relatively weaker in such a narrowly defined context.

\begin{figure}[ht]
\centering
\includegraphics[width=0.45\textwidth]{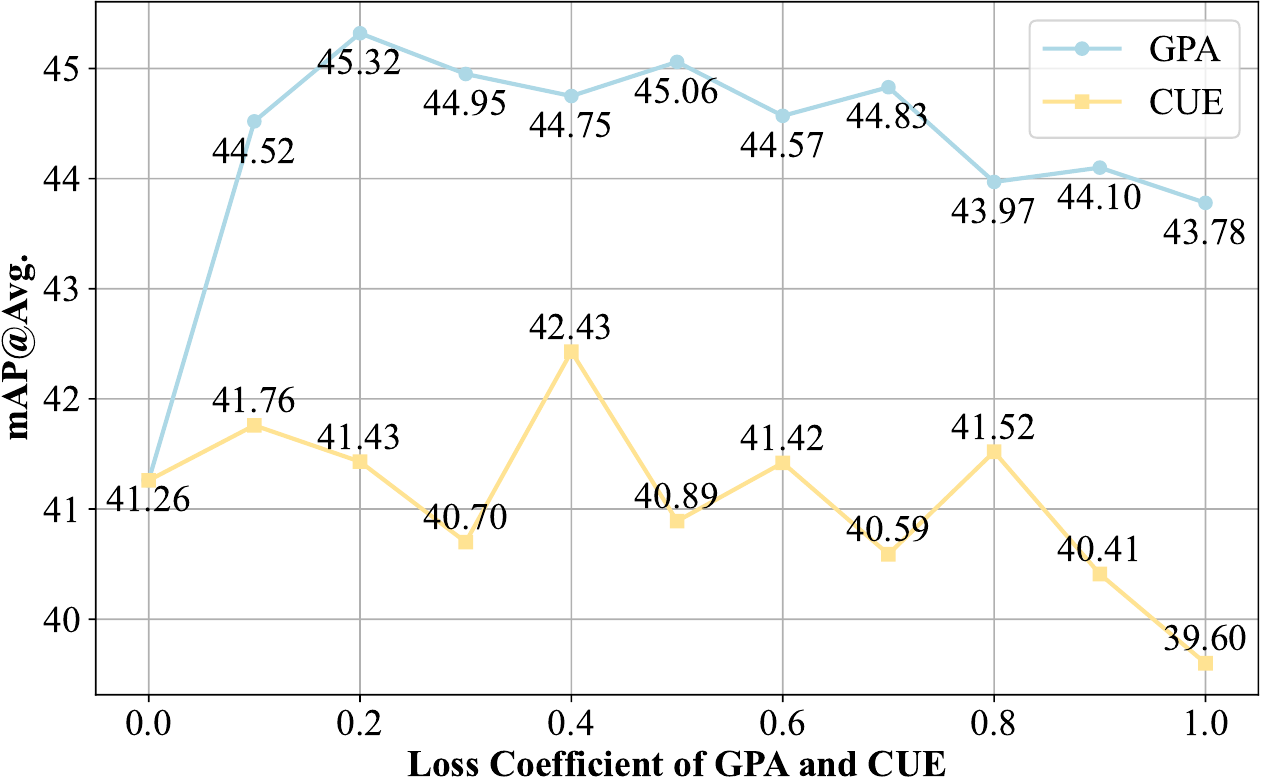} 
\caption{The impact of global partial aligner loss coefficient $\lambda_{\text{GPA}}$ and contextual understanding enhancement loss coefficient $\lambda_w$ on QVHighlights \textit{val} set without audio features.}
\label{fig4}
\end{figure}

\subsection{Visualization}
In Fig. \ref{fig3}, we visually compare the ground truth with the prediction of three leading methods and our approach on the QVHighlights \textit{val} set. In the given query ``Two TV show hosts sit on a red couch and discuss the news'', our model demonstrates a superior understanding of the keywords ``Two TV show hosts'' and ``red couch'', providing highly accurate predictions. Similarly, for the query ``Two buddies hang out above the cloudline after walking up to a viewpoint on a mountain'', our model better comprehends the critical part ``Two buddies hang out'' rather than ``a viewpoint on a mountain''. The visualization of the results proves that our model has a better understanding of the queries.

\subsection{Ablation Study}
We conduct extensive ablation experiments to demonstrate the effectiveness of the introduced modules, and the results are shown in Table \ref{table4}. Settings (b) to (d) show that each component significantly improves performance compared to the baseline (a). Although (e) does not show improvement on all evaluation metrics, the overall QDM (f), which includes QE (d) and CUE (e), achieves a significant improvement compared to (a). Settings (g) and (h) demonstrate that video enhancement and query debiasing have significant performance improvements under the guidance of GPA. The performance decrease in setting (i) indicates that without GPA, video enhancement and query debiasing do not have a better optimization direction. Setting (j) shows that, under the guidance of GPA for video and text alignment, both video enhancement and query debiasing achieve a better understanding of the video and query semantics, resulting in the best performance. As shown in Fig. \ref{fig4}, we explore the impact of the weights of $\mathcal{L}_{\text{GPA}}$ and $\mathcal{L}_w$ separately. When analyzing $\mathcal{L}_{\text{GPA}}$, the weight $\lambda_w$ is set to 0, and vice versa. We find that the model performs better when $\lambda_{\text{GPA}}$ is set to 0.2 and $\lambda_w$ is set to 0.4, respectively.

\section{Conclusion}
In this paper, we propose QD-VMR, an efficient scheme to address the deep-rooted problem of ambiguity in query semantics under the VMR task. We conduct extensive comparison and ablation experiments to verify the effectiveness of the proposed scheme, and the results show that QD-VMR achieves SOTA results on multiple metrics. Moreover, the visualization of the experimental results demonstrates that our method enhances the model’s ability to understand the contextual semantics of queries and accurately identify relevant moments.  In the future, we also hope to explore potential methods to utilize unlabeled or weakly labeled data, as well as large-scale multimodal models with abundant prior knowledge to better perform downstream tasks. We believe that the issues and methods proposed in this work can provide valuable insights into related fields.

\bibliography{aaai25}

\end{document}